%% file: root.tex
\documentclass[letterpaper, 10 pt, conference]{ieeeconf}  

\usepackage{graphicx}
\usepackage{multirow}

\usepackage{cite}
\usepackage{amsmath,amssymb,amsfonts}
\usepackage{algorithmic}
\usepackage{textcomp}
\usepackage{xcolor}
\usepackage{booktabs}

\usepackage{caption}    
\usepackage{subcaption}

\IEEEoverridecommandlockouts                              

\title{\LARGE \bf
Synthetic Enclosed Echoes: A New Dataset to Mitigate the Gap Between Simulated and Real-World Sonar Data
}

\author{Guilherme de Oliveira$^{1}$, Matheus M. dos Santos$^{2}$ and Paulo L. J. Drews-Jr$^{1}$
\thanks{*This work was financed by the Human Resource Program of The Brazilian National Agency for Petroleum, Natural Gas, and Biofuels – PRH-ANP, FINEP and CNPq.}
\thanks{$^{1}$NAUTEC - Intelligent robotics and automation group, FURG - Federal University of Rio Grande, Rio Grande, RS, Brazil
{\tt\small guilherme@furg.br}}%
\thanks{$^{2}$ICMC - Institute of Mathematics and Computer Sciences, USP - University of São Paulo, São Carlos, SP, Brazil}}

\begin{document}

\maketitle
\thispagestyle{empty}
\pagestyle{empty}

\begin{abstract}
\input{text/0_abstract}
\end{abstract}

\section{INTRODUCTION}
\input{text/1_introduction}
\section{RELATED WORK} \label{sec:lit_review}
\input{text/2_literature_review}
\section{METHODOLOGY} \label{sec:methodology}
\input{text/3.0.1_SimulatorPLUSmethodology.tex}
\input{text/3.1_methodology}
\section{RESULTS} \label{sec:results}
\input{text/4_results}
\section{CONCLUSIONS} \label{sec:conclusions}
\input{text/5_conclusion}

\section*{ACKNOWLEDGMENT}

This study was financed by the Human Resource Program of The Brazilian National Agency for Petroleum, Natural Gas, and Biofuels – PRH-ANP, supported with resources from oil companies considering the contract clause nº 50/2015 of R, D\&I of the ANP. CAPES, CNPq, FINEP, and the Intelligent robotics and automation group at FURG also support this study. LLM-based tools were used to help in the codes and text correction.


\bibliographystyle{IEEEtran}
\bibliography{bibliografia}


\end{document}

%% file: text/0_abstract.tex
This paper introduces Synthetic Enclosed Echoes (SEE), a novel dataset designed to enhance robot perception and 3D reconstruction capabilities in underwater environments. SEE comprises high-fidelity synthetic sonar data, complemented by a smaller subset of real-world sonar data. To facilitate data acquisition, a simulated environment has been developed, enabling the generation of additional data through modifications such as the inclusion of new structures or imaging sonar configurations. This hybrid approach leverages the advantages of synthetic data, including readily available ground truth and the ability to generate diverse datasets, while bridging the simulation-to-reality gap with real-world data acquired in a similar environment. The SEE dataset was used to evaluate sonar 3D reconstruction methods. These techniques were employed to validate the dataset, confirming its suitability for underwater 3D reconstruction. Furthermore, this paper proposes a novel state-of-the-art algorithm that demonstrates improved performance compared to existing methods. 

%% file: text/1_introduction.tex
3D data is valuable for various applications, including map generation, inspection tasks, and environmental monitoring~\cite{maurell2022volume}. The 3D reconstruction process involves sensors to capture environmental and target characteristics, followed by algorithms to organize and interpret the acquired data. Thereby, reconstructing the scene and target features~\cite{massot2015optical}.

Significant advancements have been made in underwater 3D reconstruction using optical sensors~\cite{hu2023overview}. However, optical sensors face inherent limitations in underwater environments, including rapid light-wave energy attenuation, high turbidity, and low-light conditions.
Consequently, these limitations often hinder the ability of optical sensing to meet the demands of real-world applications.
In contrast, acoustic sensors provide a compelling alternative for practical underwater applications. Low loss, strong diffraction, extended propagation distances, and minimal sensitivity to water quality characterize the propagation of sonar waves in water. This enables acoustic sensors to perform more effectively in complex, low-light, underwater environments compared to optical sensors~\cite{santos2019}.
\begin{figure}[!htb]
    \centering
    \includegraphics[width=0.47\textwidth]{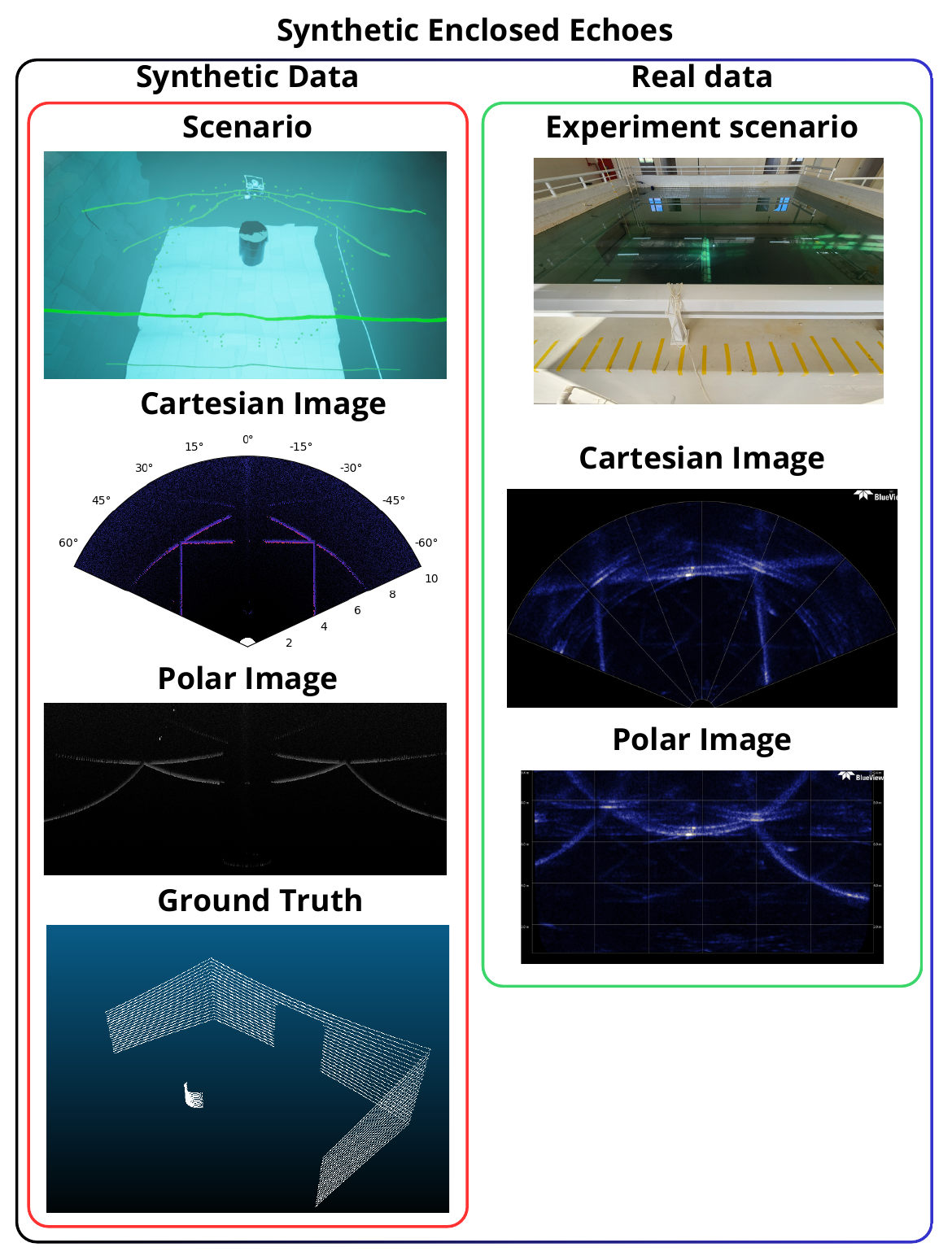}
    \caption{Synthetic Enclosed Echoes: a new dataset of synthetic and real-world sonar data for a closed underwater environment}
    \label{fig:SEE_intro}
\end{figure}

A primary challenge in processing sonar images from active acoustic sensors, such as imaging sonars, is mitigating the sensor's inherent effects, including ambiguity, reverberation, and various forms of noise~\cite{SonarTheory}. These sensor-specific issues can be partially addressed through computational strategies, such as filtering and post-processing the acquired data.

Characterizing sonar noise is a complex task, and even when modeled, it imposes significant computational overhead~\cite{ribeiro2017}. Furthermore, modeling certain effects, such as the sensor's ambiguity, presents considerable difficulties. An alternative approach to noise mitigation involves employing learning-based methods, which effectively model and abstract these intricate effects. However, a major limitation of learning algorithms is their substantial data requirement. Existing sonar datasets are notably scarce, primarily due to the high cost of sonar equipment and the logistical and infrastructural complexities associated with data collection.

Nevertheless, synthetic data for training learning methods is gaining prominence, as demonstrated in~\cite{depth_anything_v2}, where a purely synthetic dataset was used for monocular camera depth estimation. This concept of synthetic image data can be extended to sonar data, provided a reliable sonar simulation is available to accurately replicate the effects and nuances observed in real acoustic images.

Given the challenges above, this paper introduces Synthetic Enclosed Echoes (SEE), a novel sonar dataset primarily composed of synthetic acoustic images, complemented by real-world data. All data is acquired within a controlled underwater environment, specifically a tank, and the simulated environment is designed to replicate the real-world setup, minimizing the simulation-to-reality gap for sonar data. Figure \ref{fig:SEE_intro} illustrates the proposed dataset. The illustration highlights the parts of SEE, where the synthetic component comprises simulator environments, sonar images, sonar ground truth, and CAD models of the sonar-scanned objects. Additionally, the real-world component consists of Cartesian and polar sonar images captured in an environment that mirrors the synthetic simulation.

The organization of this work is as follows: Section \ref{sec:lit_review} provides a literature review of relevant works focusing on datasets for sonar perception and 3D reconstruction. Section \ref{sec:methodology} presents the proposed methodology, detailing the stages and design decisions involved in developing the SEE dataset, and shows our modifications to a state-of-the-art algorithm for 3D reconstruction of underwater environments using sonar data. Section \ref{sec:results} presents the results and experiments conducted using the SEE dataset, and Section \ref{sec:conclusions} concludes the paper.

%% file: text/2_literature_review.tex
The challenge of 3D reconstruction from sonar images remains a subject of ongoing research, primarily due to the inherent noise and ambiguities in sonar data. Forward-Looking Sonars (FLS) are considered a promising modality for 3D imaging and reconstruction among various sonar types. The existing literature on 3D reconstruction using sonar images can be broadly categorized into two main approaches: classical, mathematically based methods and learning-based methods.

Within the domain of classical methods, the work of \cite{guerneve2015underwater} stands out, employing a BlueView imaging sonar for 3D reconstruction. Their proposed methodology reconstructs targets without requiring prior knowledge of their shape. This approach utilizes a simplified mathematical model of the sonar, coupled with filtering techniques, to achieve reconstruction from a sequence of acquired images. Other studies, such as \cite{rahman2019contour}, propose a system incorporating a stereo camera, an inertial sensor, a depth sensor, and a mechanical scanning profiling sonar to perform contour-based reconstruction.

McConnell et al. \cite{mcconnell2020fusing} proposed a novel strategy to mitigate the ambiguity associated with the sonar's elevation angle. Their approach involves employing two FLS sonars, one positioned vertically and the other horizontally. This configuration generates overlapping fields of view, facilitating point matching and correcting ambiguity errors. Kim et al. \cite{kim2021high} proposed a methodology for generating 3D maps from a single sonar mounted on an autonomous underwater vehicle (AUV). Their proposal relies on two primary processes: 2D image reconstruction and point cloud generation. The fusion of these two methods produces a 3D map.

The application of learning-based methods is gaining increasing prominence in the field of 3D reconstruction for sonar imaging. These methods can be broadly categorized into supervised and self-supervised learning. The key distinction between these approaches lies in the use of ground truth data. Supervised methods leverage labeled data, while self-supervised methods do not. Within supervised learning, Sung et al. \cite{sung2020underwater} proposed a sonar-based underwater object classification method by reconstructing an object's 3D geometry. In their work, a point cloud is generated from sonar images, and a neural network is subsequently employed to predict the object's class based on the generated point cloud.

Debortoli et al. \cite{debortoli2019elevatenet} propose an application of a neural network to address the ambiguity problem in sonar images. Their strategy employs a modified U-Net architecture~\cite{ronneberger2015u}, utilizing an elevation map as ground truth. Debortoli in this work leverages synthetic data to create a small dataset, with the network initially trained on this synthetic data generated using a simulator developed by Cerqueira et al. \cite{cerqueira2017novel}. The synthetic dataset comprises 8,454 images of spheres, cylinders, and cubes. Subsequent fine-tuning is performed using real-world data, consisting of 4,667 sonar images obtained with a Tritech Gemini 720i imaging sonar from custom-built concrete targets in the shape of spheres, cylinders, and cubes.

Wang et al. \cite{wang2022learning, wang2021elevation} proposed a novel ground truth representation that generates a pseudo-front-depth map. This approach aims to perform 3D reconstruction through supervised learning. They created a small simulated dataset using a Blender-based simulator. The dataset comprises an artificial environment, terrain, and spheres, with 3D CAD models serving as ground truth. The real-world dataset was acquired in a large-scale water tank using a Sound Metrics ARIS EXPLORER 3000 imaging sonar, replicating the simulated objects and scenarios.

Within the domain of self-supervised learning, Qadri et al. \cite{qadri2023neural} presented a technique for 3D reconstruction based on a neural network architecture named NEUSIS. Their work involved creating both simulated and real-world datasets. The simulated dataset, acquired using HoloOcean \cite{potokar2022holoocean}, comprises sonar images of two objects, each in three distinct configurations. The real-world datasets were collected using a Sound Metrics DIDSON imaging sonar to capture data from a test structure submerged in a test tank. Three datasets were captured, corresponding to the sonar's feasible elevation apertures (1°, 14°, and 28°). Their approach was subsequently compared with other reconstruction algorithms designed for acoustic images, yielding highly satisfactory results and demonstrating superior performance without incurring their associated memory overhead.

Wang et al. \cite{wang2023motion} employed a self-supervised learning methodology, utilizing the same dataset as in \cite{wang2022learning} and \cite{wang2021elevation}. However, sensor position and motion information were incorporated as inputs for each sonar image, enabling self-supervised learning. The methodology was validated using both synthetic and real-world data.

In summary, while significant strides have been made in employing classical and learning-based methods for sonar image processing, the literature reveals a notable disparity in dataset availability. As highlighted by \cite{aubard2024sonar} in their survey, most publicly available sonar datasets are primarily geared towards object classification tasks. Conversely, datasets specifically designed for 3D reconstruction from sonar imagery remain severely limited. This scarcity underscores the need for dedicated datasets, such as the SEE dataset proposed in this work, which is more complete by providing a greater variety of objects and more detailed feature information, to facilitate advancements in 3D reconstruction methodologies for underwater environments.

%% file: text/3.0.1_SimulatorPLUSmethodology.tex
For this study, HoloOcean~\cite{potokar2022holoocean}~\cite{potokar2022holooceanSonar} was selected as the simulation platform. Several factors influenced this decision, including the high fidelity of its sonar imaging simulations, the ongoing updates and development of the simulator, and its extensive expandability as an open-source project. Another factor is its implementation within the Unreal Engine, which enables the creation of high-quality visual effects and large-scale scenarios with manageable computational demands. The primary drawback associated with HoloOcean's sonar simulation is its potential computational cost, which can vary depending on the resolution and noise levels applied.

\begin{figure}[!htb] 
\centering
\includegraphics[width=0.4\textwidth]{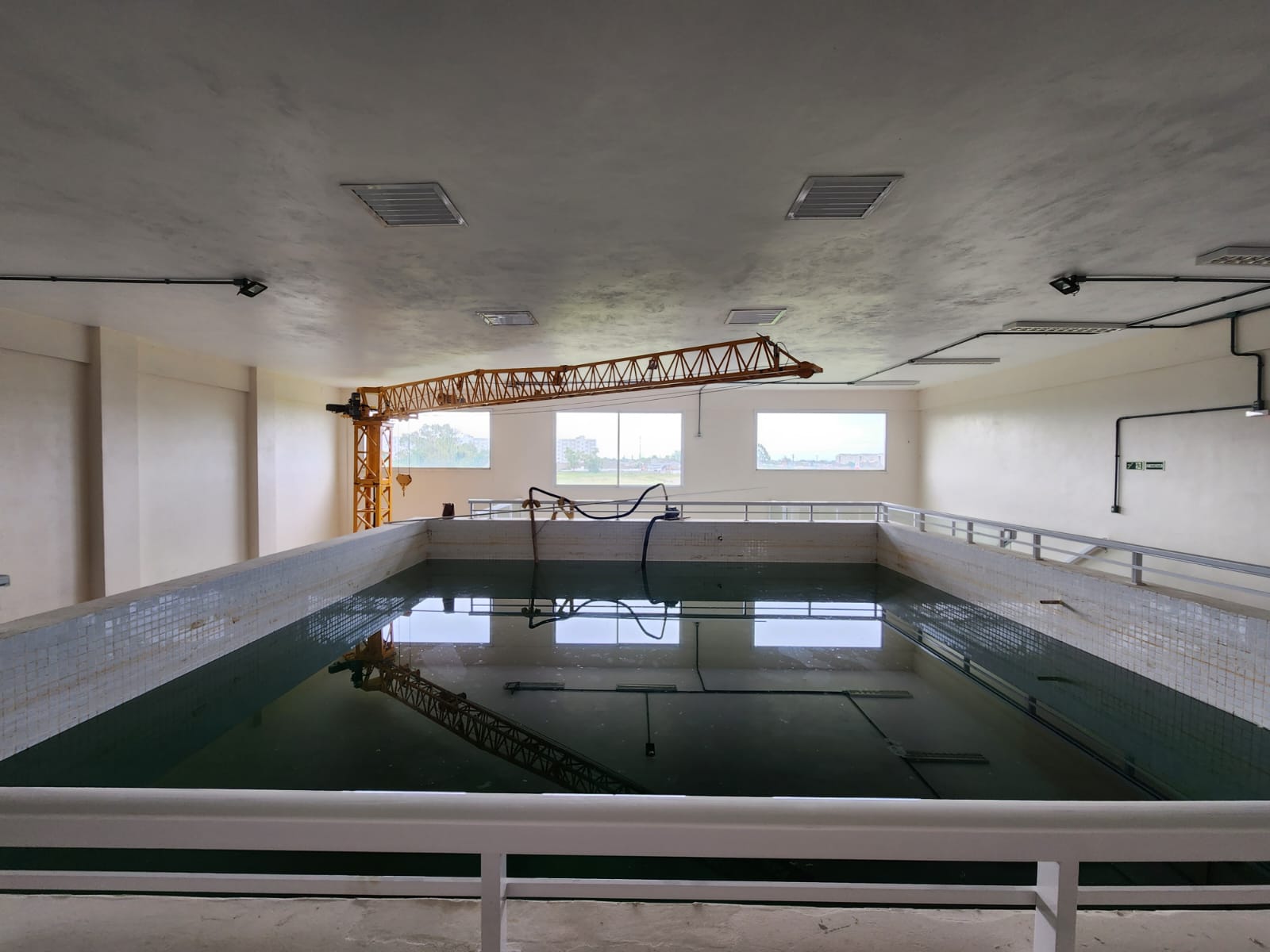}
\caption{\label{fig:aquatec} Indoor Tank Facilities.}
\end{figure}

Simulation was employed to facilitate the development of transfer strategies from simulated to real-world settings. This approach leverages the ease of generating highly reliable synthetic data, thereby minimizing the need for extensive real-world missions and experiments, which leads to significant cost reductions and accelerated development cycles. To this end, a simulated environment was created to replicate an indoor tank facility, as depicted in Figure \ref{fig:aquatec}. The tank's dimensions, 7 meters in width, 7 meters in length, and 5 meters in depth, along with its constituent materials, were modeled within the simulator.

%% file: text/3.1_methodology.tex
\begin{figure*}[!htb] 
    \centering
    \includegraphics[width=0.8\textwidth]{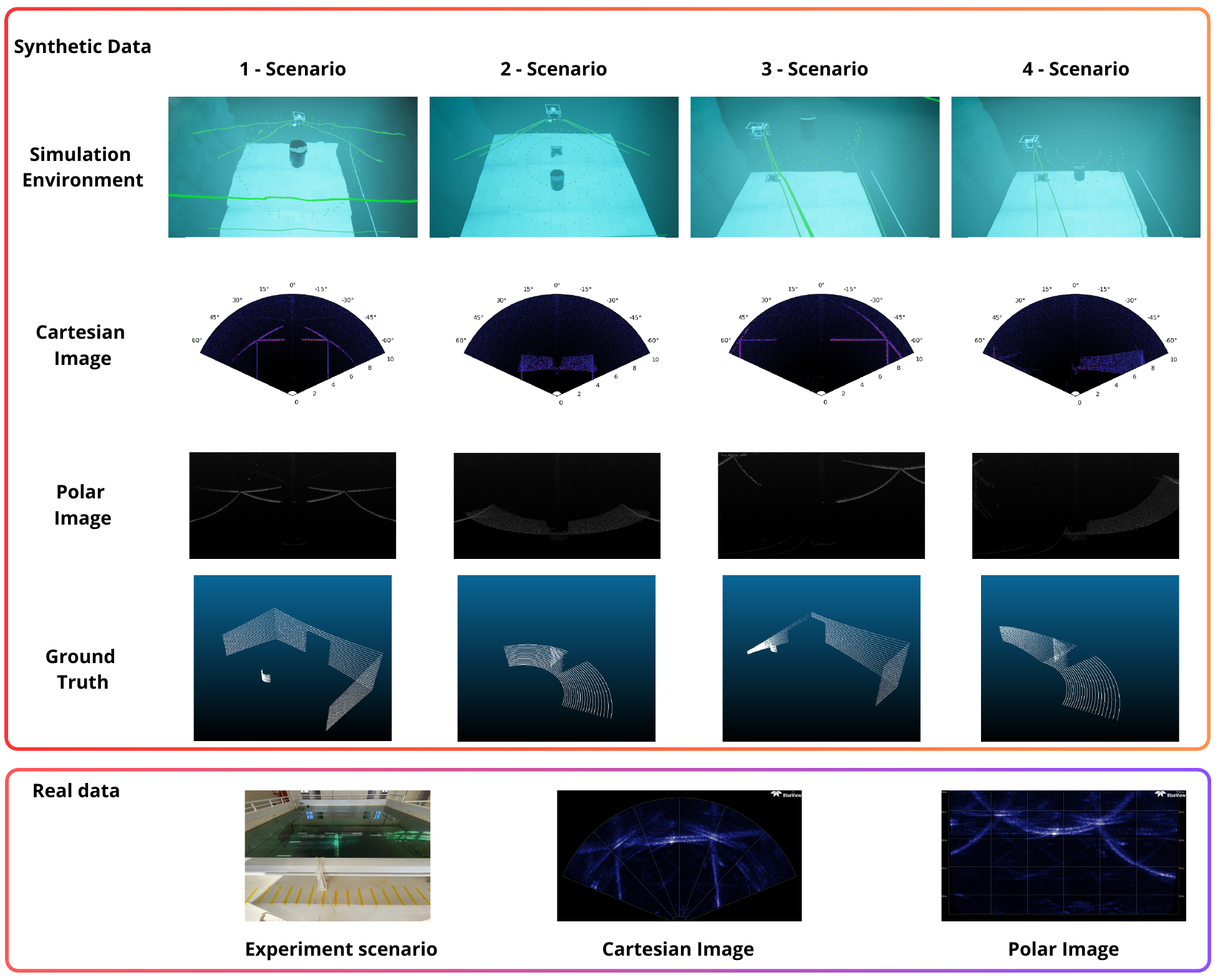}
    \caption{An overview of the SEE structure, exemplifying the scenarios developed and their respective collected data.}
    \label{fig:SEE_main}
\end{figure*}

Having established the simulator and the reference environment, this study necessitates a dataset with associated ground truth information to serve as a training foundation for learning-based methods. To this end, a set of key criteria was defined to guide the construction and generation of this dataset:

\begin{itemize}
    \item The dataset must comprise sonar data with corresponding ground truth.
    \item The simulation environment must be expandable and replicable.
    \item The simulation environment must be capable of simulating diverse sonar configurations.
    \item The environment must feature a range of objects, from simple to complex structures, relevant to underwater environments.
    \item The simulation environment must provide a variety of scenarios within a closed setting.
    \item Data collection must be structured, including positional reference information for each image and multi-angle image acquisition.
\end{itemize}

\subsection{Dataset Development}
A virtual world was constructed using the Unreal Editor to meet the dataset requirements, replicating the physical tank. A diverse array of objects was incorporated, including geometric solids, helices, anchors, and underwater civil construction materials such as trusses and concrete blocks. The object dimensions were carefully selected to ensure they could be positioned within the tank, allowing an autonomous underwater vehicle (AUV) to navigate and collect data without collisions.


Due to simulator limitations, specifically the inability to dynamically position objects for sonar interpretation, developing a static world with pre-positioned objects was necessary. Consequently, four distinct worlds were created, each comprising 64 tanks, with 40 tanks in each scenario populated with simulation props; a simulation prop is a virtual object used within a simulation to represent a real-world item or element. The scenarios differed in prop placement, with the 1st scenario featuring floating props, the 2nd scenario featuring bottom-positioned props, the 3rd scenario featuring wall-proximal floating props, and the 4th scenario featuring wall-proximal bottom-positioned props. This design aimed to increase the variability in object location, ranging from simple to challenging scenarios, thereby enhancing the diversity of the simulated data. Figure \ref{fig:SEE_main} illustrates the developed worlds and their respective scenarios.

\begin{figure*}[!htb] 
    \centering
    \includegraphics[width=0.8\textwidth]{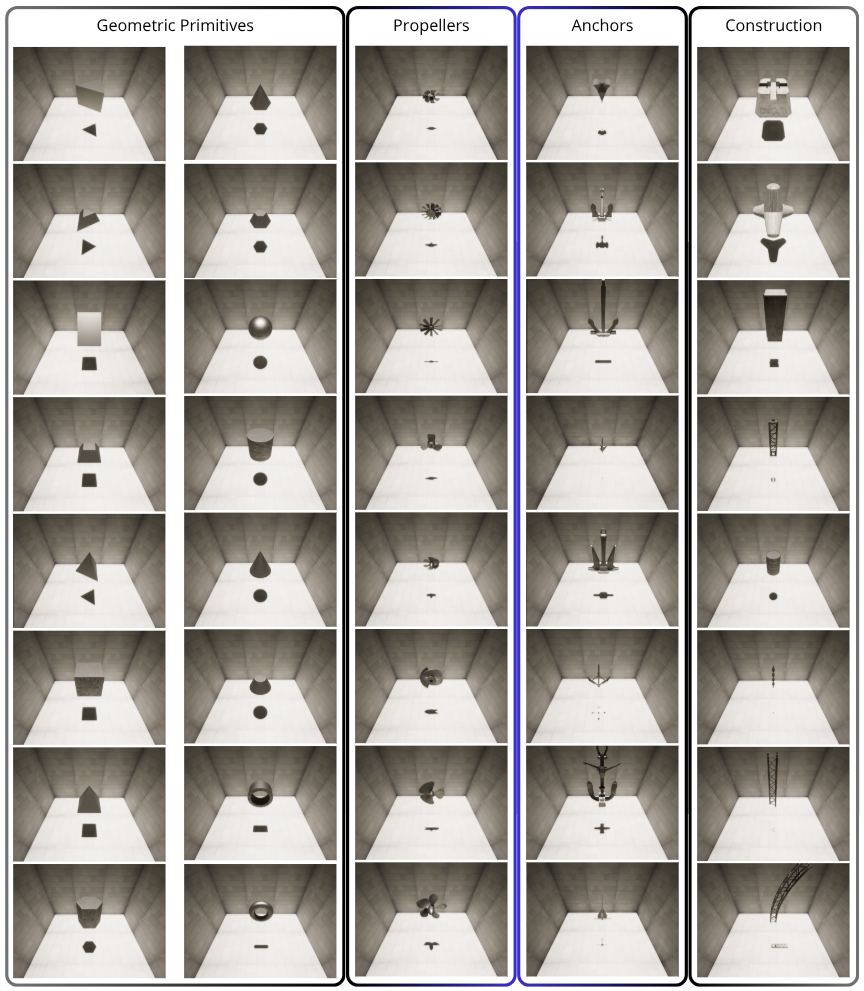}
    \caption{An overview of all props present in the SEE dataset.}
    \label{fig:SEE_assets}
\end{figure*}       

The dataset proposed in this work is based on structures typically found in underwater scenarios, such as barrels, trusses, and other structures widely used in submerged constructions. Figure \ref{fig:SEE_assets} exemplifies all assets used in the dataset. These objects' Computer-Aided Design (CAD) files serve as ground truth. The same objects are instantiated within the simulated environment with their geometry and materials duly described. The idea is that each CAD file has an associated point cloud and sonar images with different noise intensities.

The dataset proposed in this study features structures commonly encountered in underwater scenarios, including barrels, trusses, and other assets prevalent in submerged construction, as illustrated in Figure \ref{fig:SEE_assets}. Computer-Aided Design (CAD) files for these objects serve as the foundational geometric ground truth. An array of simulated rangefinders was employed to generate accurate sonar ground truth information, positioned at theta and phi angles equivalent to the sonar used for data collection. These rangefinders provide precise distance measurements, enabling the creation of reliable ground truth data that aligns with the sonar's perspective. The output of this rangefinder array can be either a point cloud or a sonar image with an elevation angle label. The same objects are instantiated within the simulated environment, with their geometry and material properties represented. Each CAD file is associated with a corresponding point cloud and sonar images exhibiting varying noise intensities.

\begin{figure}[!htb] 
\centering
\includegraphics[width=0.45\textwidth]{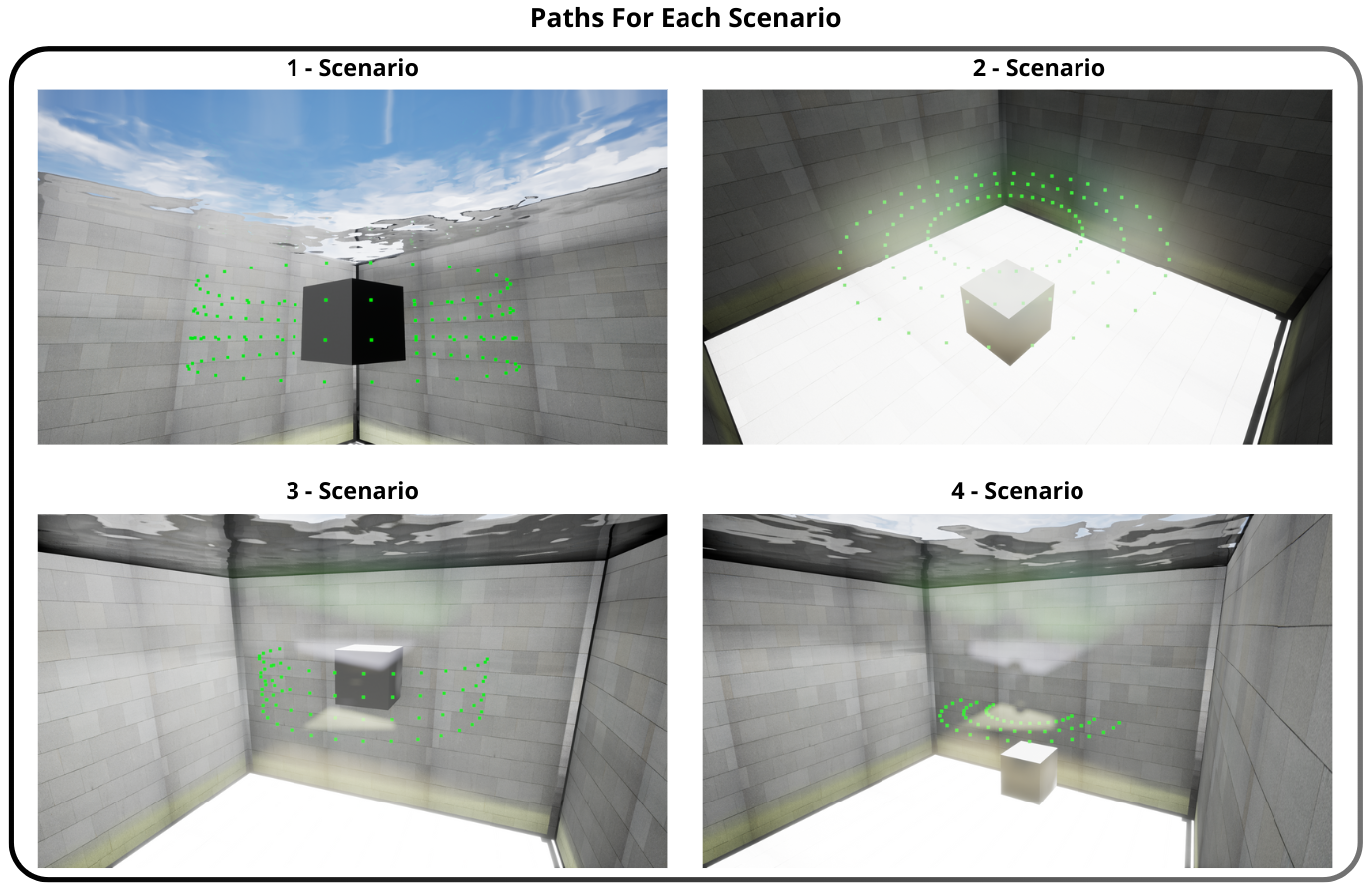}
\caption{\label{fig:path} Proposed scenarios for the dataset and their respective way-points}
\end{figure}

Upon completion of the simulation worlds, a standardized and replicable data collection methodology was established to ensure consistency across the various scenarios. The adopted strategy involved a circular trajectory around the target, with a radius of $2$ meters and waypoints defined at $10^\circ$ intervals. The Vehicle was programmed to maintain its center aligned with the object, ensuring continuous coverage within the sonar's field of view. This circular trajectory was repeated along the object's length, with a $0.3$ meter gap between each trajectory. Figure \ref{fig:path} illustrates the trajectory employed for standardized data acquisition.

For the first scenario, the object floated and was centered within the tank. Data collection was performed using a ring path, where the Vehicle's height was varied upon completion of each lap around the object. This circular path had a radius of $2$ meters, and subsequent circles were created and spaced $0.3$ meters apart until the object's entire height was covered.

In the second scenario, the object is fixed to the tank's bottom. The trajectory is circular, maintaining a constant height while varying the radius of the circle. The reference points are consistently positioned $1$ meter above the object, and the sonar is rotated by $45$ degrees.

The third scenario mirrors the first, with the distinction that the object is situated near a tank wall. A semi-circular trajectory, similar to the first scenario, is employed. The fourth scenario combines elements of the second and third scenarios. The object is positioned at the bottom of the tank and near a wall, creating a challenging 3D reconstruction scenario. Data collection follows the pattern established in the second scenario, but with a semi-circular trajectory. Figure \ref{fig:path} summarizes the four projected scenarios and their respective paths, with green points indicating the waypoints traversed by the Vehicle during data acquisition.

A modified BlueROV2 Heavy, configured with 6 degrees of freedom (DoF), was employed for real-world data acquisition. This Vehicle was adapted to carry the sonar and replicate the motion capabilities of the simulated Vehicle, which features eight thrusters and six DoF. The objective was to execute a trajectory closely resembling the simulated one, thereby minimizing the gap between simulation and reality.


The BlueView P900 sonar was employed for real-world sonar image acquisition. All simulations were designed to accurately represent this sensor, preserving its characteristics and technical specifications, including wave frequency and resolution. This approach aimed to generate simulated data closely aligned with real-world measurements. Furthermore, the simulation environment is designed to be adaptable, allowing for the representation of other sonar models and resolutions to accommodate future research demands.

The resulting dataset from the sonar data collection comprises simulations of the BlueView P900, along with associated metadata for each image. This metadata includes Vehicle and sonar position and orientation, sonar characteristics such as field of view and range, noise settings, and image timestamps. This information is only present in the synthetic data.


In addition to position data and sonar images, the dataset comprises Cartesian sonar images, polar sonar images, and ground truth data in the form of point clouds, which describe the 3D field of view of the sonar. This structure aligns with the ground truth model proposed by Debortoli et al. \cite{debortoli2019elevatenet}. Furthermore, the dataset includes structured data compatible with the methodology proposed by Qadri et al. \cite{qadri2023neural}. The dataset comprises 15,536 individual sonar images, with the ground truth and metadata files increasing the total dataset size to 142,361 files, occupying 64.7 GB of storage. 

\subsection{Validation Methodology}

Sonar 3D reconstruction methodologies were implemented to compare and validate the collected data. These included a classical reconstruction method, assuming a $0^\circ$ sonar elevation angle and considering only points with an intensity equal to or greater than $95\%$ of the maximum intensity value in each image. Neusis \cite{qadri2023neural} and ElevateNET \cite{debortoli2019elevatenet} were used as comparative benchmarks.


A methodology based on ElevateNET \cite{debortoli2019elevatenet} was proposed to complement the study. This involved modifying the network architecture to function as a regression model rather than a classification model; this methodology developed in this work will be referred to as ElevateNET-R. Two distinct training approaches were employed to thoroughly evaluate the model's performance.

The first training method utilized a standard dataset split, with $90\%$ of the images allocated for training and $10\%$ for validation. It is essential to note that although the same props (i.e., the specific underwater objects being reconstructed) may appear in both the training and validation sets, the images represent different viewpoints or perspectives of these props.

The second training method evaluated the network's generalization capability to unseen objects. In this approach, called ElevateNET-R*, all data associated with specific reconstructed assets (props) were completely excluded from the training and validation set. The remaining data is split into $90\%$ 

For both training sessions, a batch size of 32 was used, with training conducted for 150 epochs and a learning rate of $1e-6$, resulting in a training time of approximately 2 hours. The hardware utilized for this training consisted of a computer with 64GB of RAM and an NVIDIA RTX 4090 GPU with 24GB of VRAM.

%% file: text/4_results.tex
A series of experiments was conducted to evaluate current 3D reconstruction methods. Three distinct objects, a cone, a propeller, and a tetrapod, were selected for this comparative analysis across all four simulation scenarios. Figure \ref{fig:result} provides a visual comparison between the reconstruction results of each method and the ground truth. For this evaluation, only the synthetic data is used.

\begin{figure*}[htbp]
    \centering
    \begin{subfigure}[t]{0.4\textwidth}
        \includegraphics[width=\linewidth]{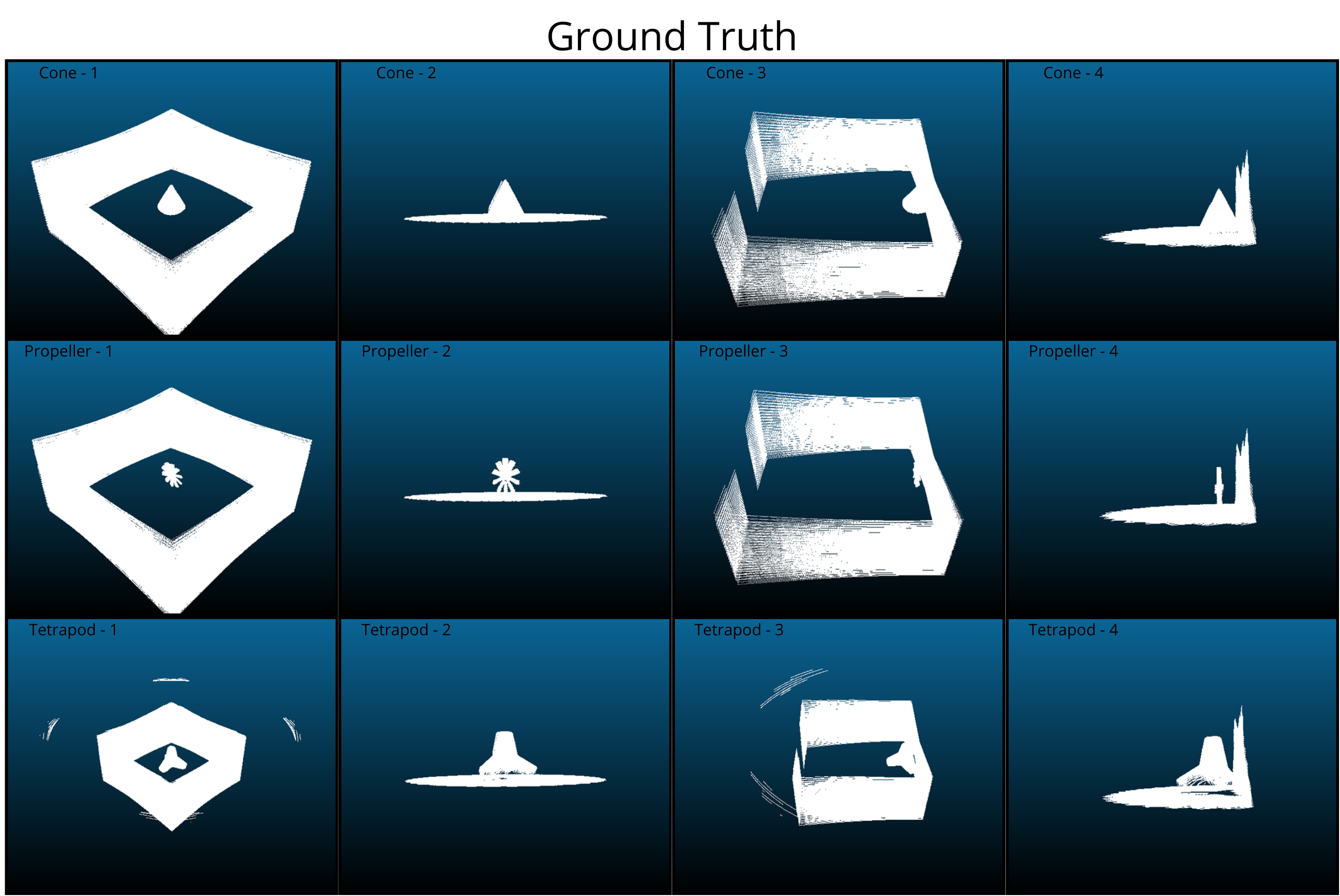}
        \subcaption{Ground Truth}
        \label{fig:result} 
    \end{subfigure}
    \begin{subfigure}[t]{0.4\textwidth}
        \includegraphics[width=\linewidth]{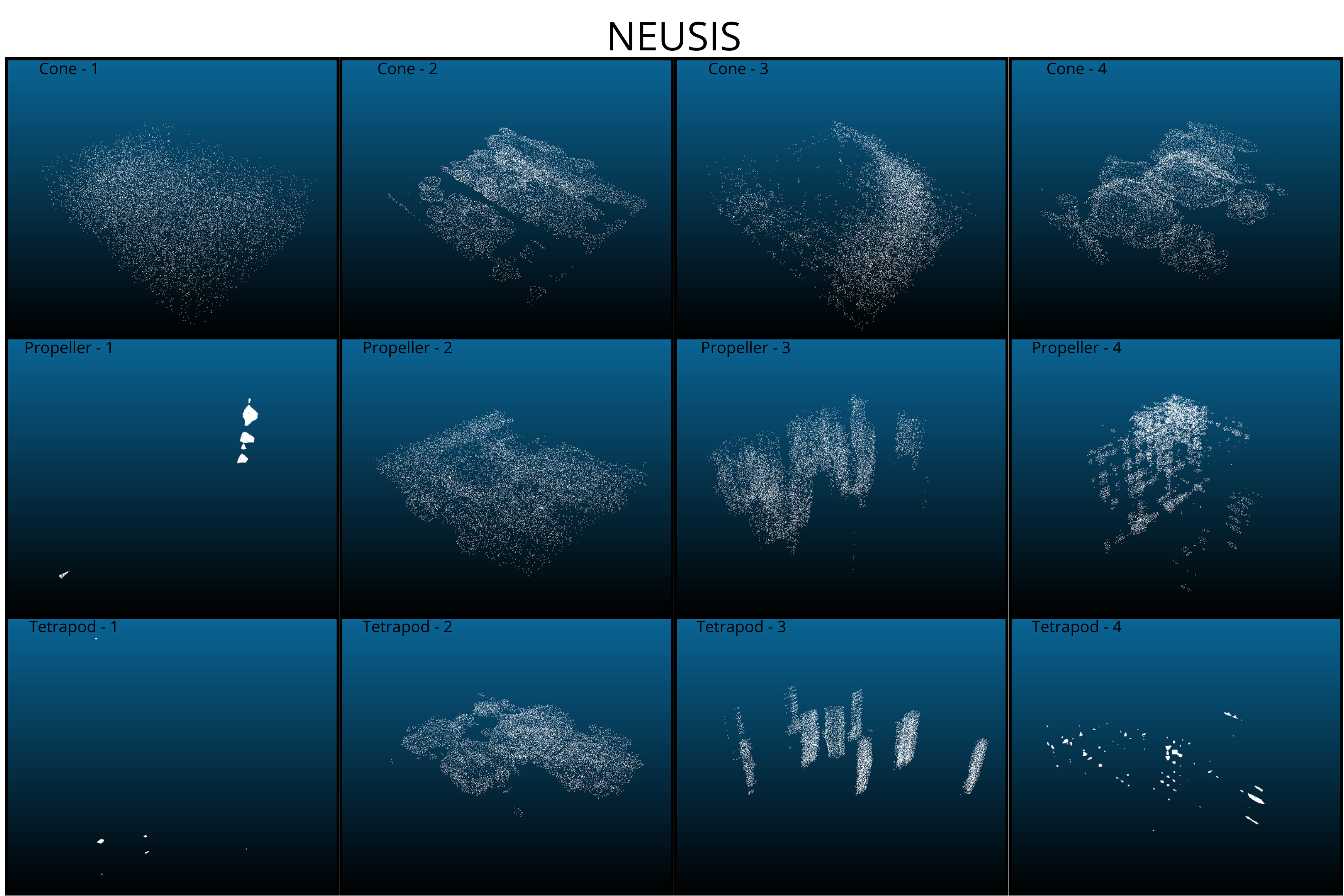}
        \subcaption{3D Reconstruction with Neusis~\cite{qadri2023neural}}
        \label{subfig:mesh}
    \end{subfigure}
    \begin{subfigure}[t]{0.4\textwidth}
        \includegraphics[width=\linewidth]{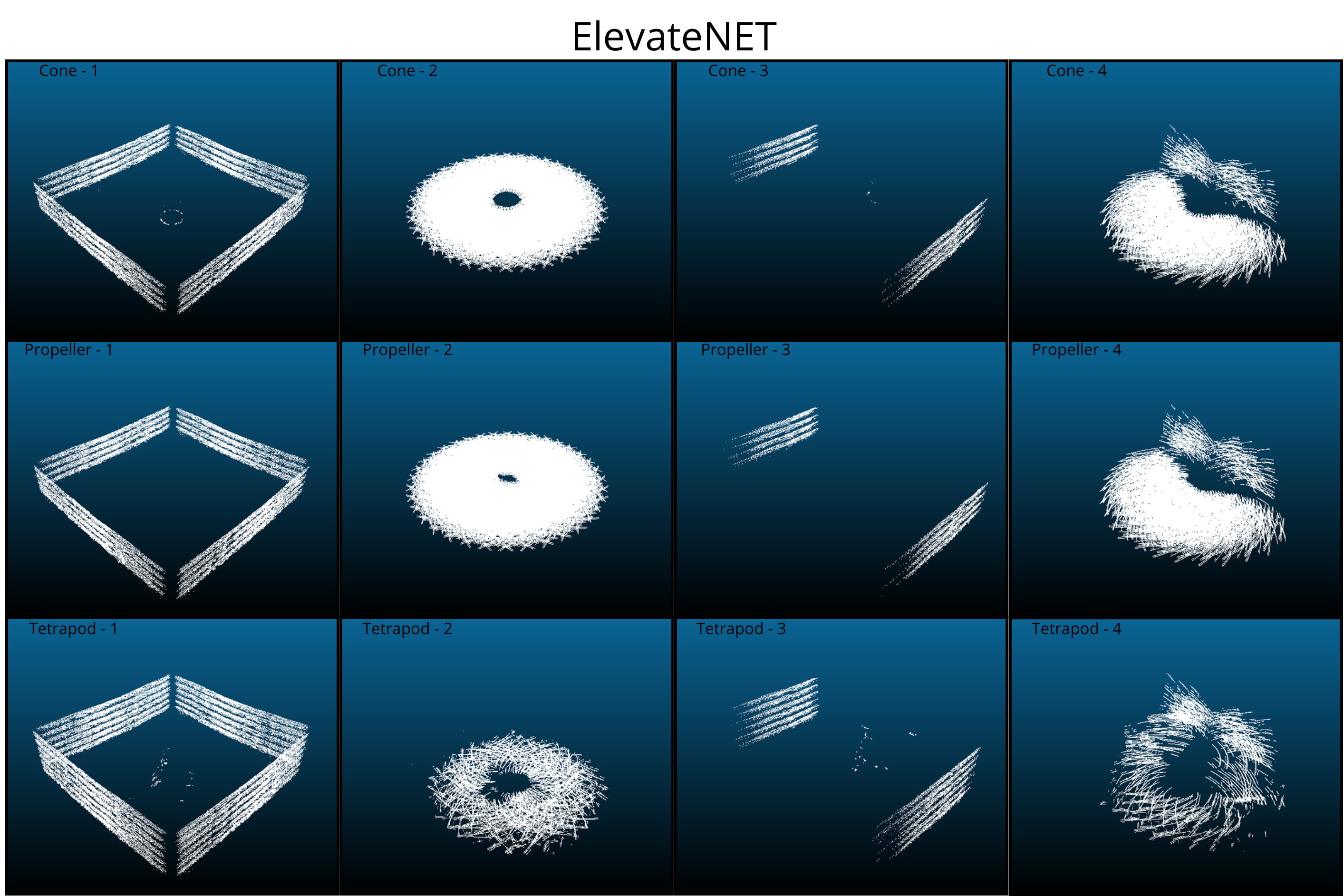}
        \subcaption{3D Reconstruction with ElevateNET~\cite{debortoli2019elevatenet}}
        \label{subfig:cylindrical}
    \end{subfigure}
    \begin{subfigure}[t]{0.4\textwidth}
        \includegraphics[width=\linewidth]{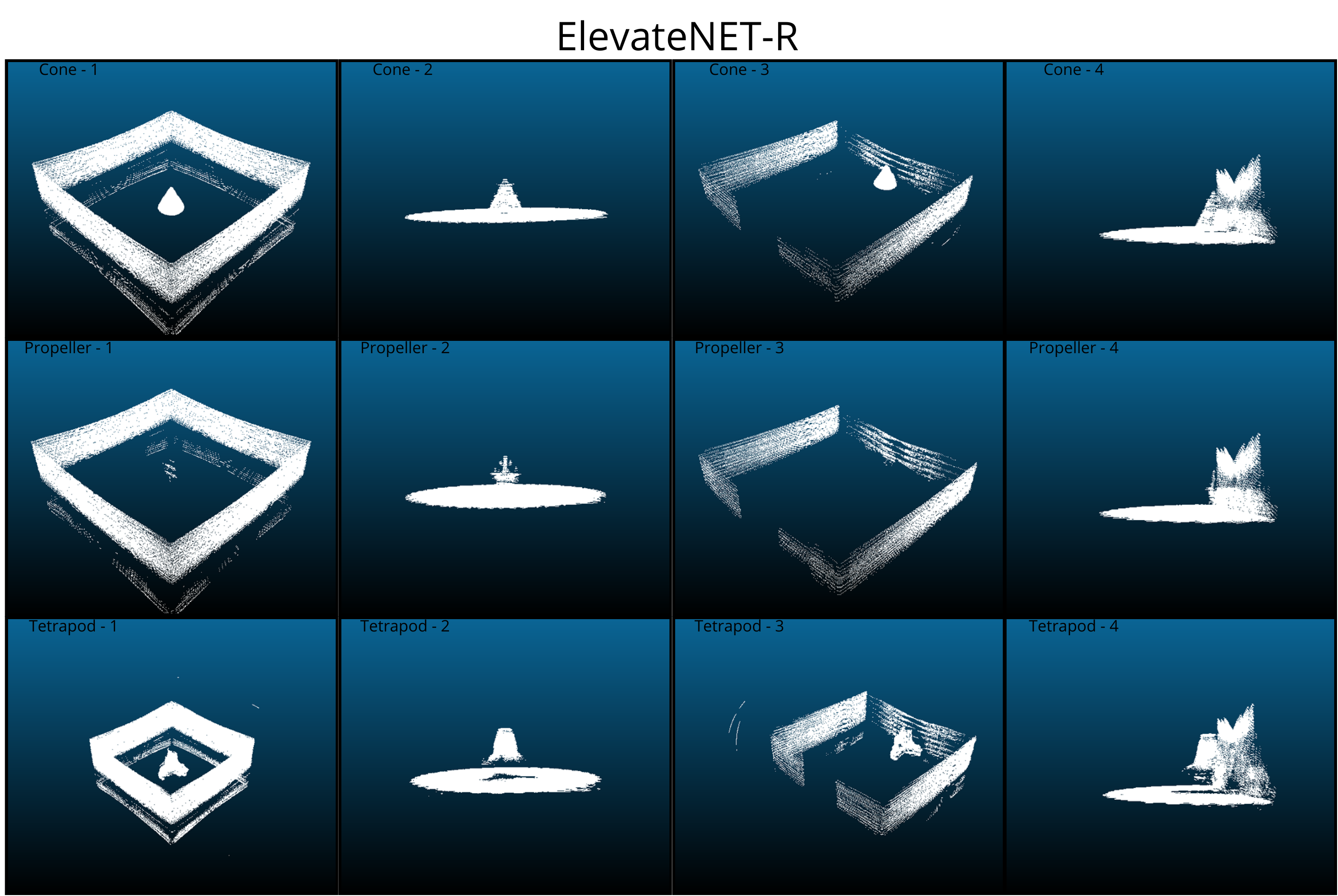}
        \subcaption{3D Reconstruction with the ElevateNET-R.}
        \label{subfig:cylindrical}
    \end{subfigure}
    \begin{subfigure}[t]{0.4\textwidth}
        \includegraphics[width=\linewidth]{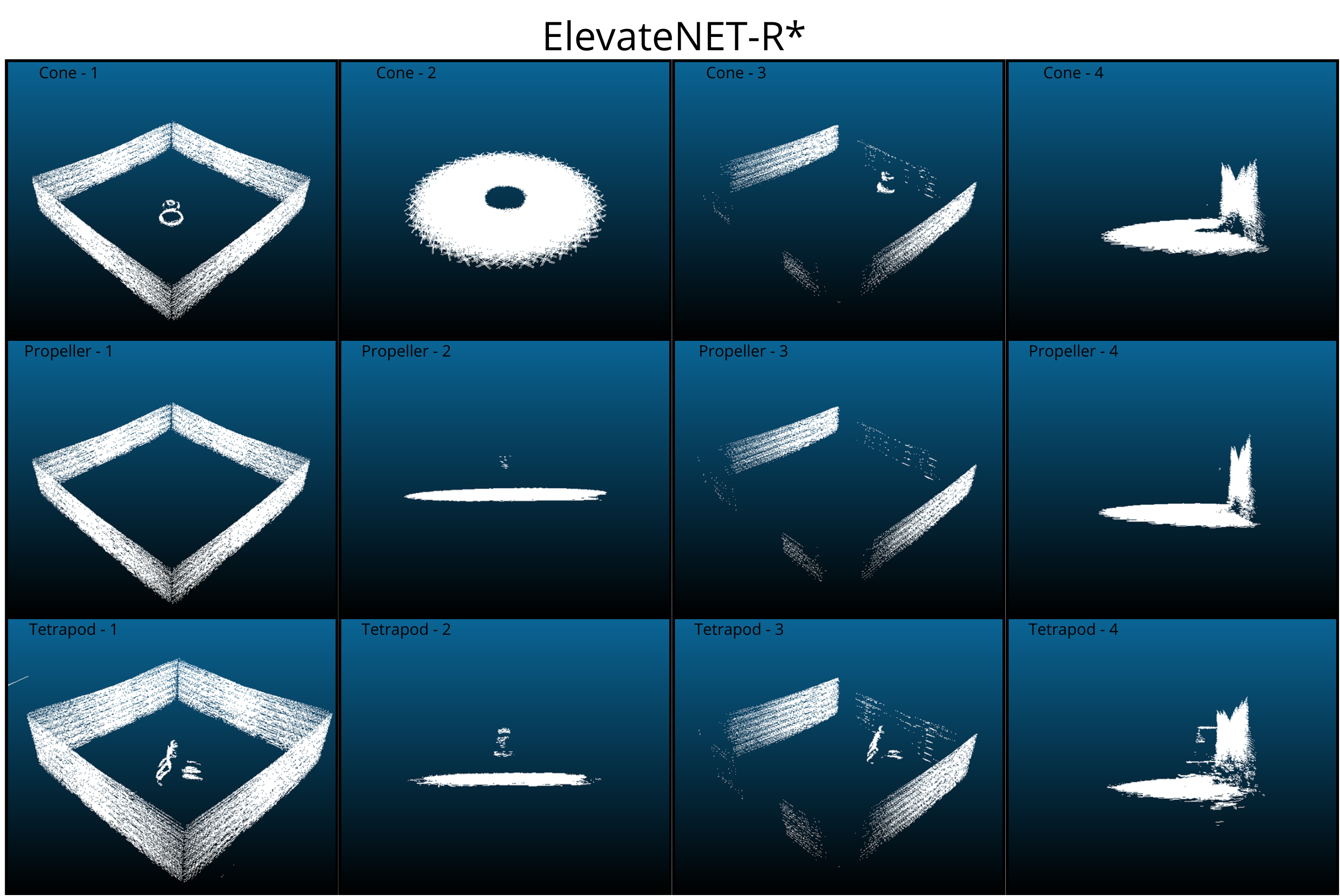}
        \subcaption{3D Reconstruction with the ElevateNET-R*.}
        \label{subfig:cylindrical}
    \end{subfigure}
    \caption{A visual comparison between the results obtained with all tested methodologies and Ground Truth.}
    \label{fig:result}
\end{figure*}

It was observed that Neusis~\cite{qadri2023neural} failed to produce any discernible 3D reconstructions. The training of this method was performed on a GPU cluster comprising two NVIDIA Tesla V100 GPUs with 32GB of VRAM, enabling the simultaneous training of two models. The training duration ranged from 20 to 35 hours. It is hypothesized that Neusis's poor performance may be attributed to its original design, which does not account for environments containing multiple objects.

The classical approach exhibited significant limitations, yielding the sparsest point clouds. ElevateNET~\cite{debortoli2019elevatenet} successfully reconstructed the tank walls and bottom but produced suboptimal results for the primary objects of interest.

The proposed methodology outperformed ElevateNET in all scenarios, achieving superior reconstruction quality. While these results could indicate overfitting, the performance of ElevateNET R*, where the reconstructed objects were excluded from the training dataset, demonstrated the proposed method's ability to learn generalized features, surpassing the original ElevateNET. The Hausdorff distance, both in terms of mean and root mean square, was employed for quantitative comparison. Table \ref{tab:result} summarizes the numerical results.

\begin{table*}[htbp]
\centering
\begin{tabular}{l|ll|ll|ll|ll}
\multicolumn{1}{c|}{\multirow{2}{*}{Objects}} &
  \multicolumn{2}{c|}{Neusis} &
  \multicolumn{2}{c|}{ElevateNET} &
  \multicolumn{2}{c|}{ElevateNET-R} &
  \multicolumn{2}{c}{ElevateNET-R*} \\
\multicolumn{1}{c|}{} &
  \multicolumn{1}{c}{Mean} &
  \multicolumn{1}{c|}{RMS} &
  \multicolumn{1}{c}{Mean} &
  \multicolumn{1}{c|}{RMS} &
  \multicolumn{1}{c}{Mean} &
  \multicolumn{1}{c|}{RMS} &
  \multicolumn{1}{c}{Mean} &
  \multicolumn{1}{c}{RMS} \\ \hline
Cone - 1      & 1.8332017 . & 2.6673460 & 0.2001159 & 0.4088612 & \textbf{0.0356741} & \textbf{0.0783030} & 0.1419549 & 0.3244358   \\ \hline
Cone - 2      & 4.9558206   & 5.2943482 & 0.0343334 & 0.1298238 & \textbf{0.0075557} & \textbf{0.0120230} & 0.0383856 & 0.1434662   \\ \hline
Cone - 3      & 3.1134795   & 4.0114953 & 0.4760223 & 0.7963873 & \textbf{0.0752777} & \textbf{0.1940996} & 0.1429360 & 0.2886642 . \\ \hline
Cone - 4      & 4.6956369   & 4.9297677 & 0.0429019 & 0.1117694 & \textbf{0.0092025} & \textbf{0.0143699} & 0.0337111 & 0.1044244   \\ \hline
Propeller -1  & 4.1635870   & 5.3039222 & 0.2886401 & 0.6733100 & \textbf{0.0599430} & \textbf{0.1385479} & 0.2320398 & 0.6221491   \\ \hline
Propeller - 2 & 5.2432549   & 5.6456313 & 0.0237024 & 0.1043392 & \textbf{0.0072436} & \textbf{0.0117776} & 0.0114937 & 0.0356924   \\ \hline
Propeller - 3 & 1.7457104   & 2.3335265 & 0.8988169 & 1.4825769 & \textbf{0.0959434} & \textbf{0.2313184} & 0.1516724 & 0.3000660   \\ \hline
Propeller - 4 & 4.6402903   & 5.2186675 & 0.0321321 & 0.0856037 & \textbf{0.0098836} & \textbf{0.0174073} & 0.0198051 & 0.0636210   \\ \hline
Tetrapod - 1&
  2.2253484 &
  2.6596196 &
  0.1524595 &
  0.3428520 &
  \textbf{0.0268286} &
  \textbf{0.0774420} &
  0.1051520 &
  0.2741911 \\ \hline
Tetrapod - 2  & 4.8643184   & 5.1939374 & 0.0813025 & 0.1790147 & \textbf{0.0119844} & \textbf{0.0288320} & 0.0553840 & 0.1078482   \\ \hline
Tetrapod - 3  & 2.8957832   & 3.8481003 & 0.3059703 & 0.5775840 & \textbf{0.0483592} & \textbf{0.1291978} & 0.1167501 & 0.2524158   \\ \hline
Tetrapod - 4  & 5.6930161   & 5.9554727 & 0.1251040 & 0.2134584 & \textbf{0.0171282} & \textbf{0.0403382} & 0.0474862 & 0.0977827   \\ \hline
\end{tabular}
\caption{Root mean square (RMS) and mean Hausdorff distance errors for the proposed experiments.}
\label{tab:result}
\end{table*}


While this result reinforces the argument that current methods are not generic enough, as evidenced by their limited validation on more diverse datasets, this work offers a step towards fostering the development of more robust and generic approaches. To the best of our knowledge, this is the first work to introduce a comprehensive dataset and a versatile simulation environment capable of generating a wide variety of high-fidelity synthetic sonar data. It is crucial to emphasize that all data utilized in this study's training and evaluation processes are exclusively synthetic, allowing for controlled experimentation and detailed analysis of model behavior under various simulated conditions. Furthermore, the SEE dataset includes a portion of real-world sonar data, which, while not the focus of the current evaluation, holds potential for future work to assess the generalization capabilities of trained models to real-world scenarios and bridge the gap between synthetic and real-world sonar perception.

\begin{figure*}[htbp]
    \centering
    \begin{subfigure}[t]{0.32\textwidth}
        \includegraphics[width=\linewidth]{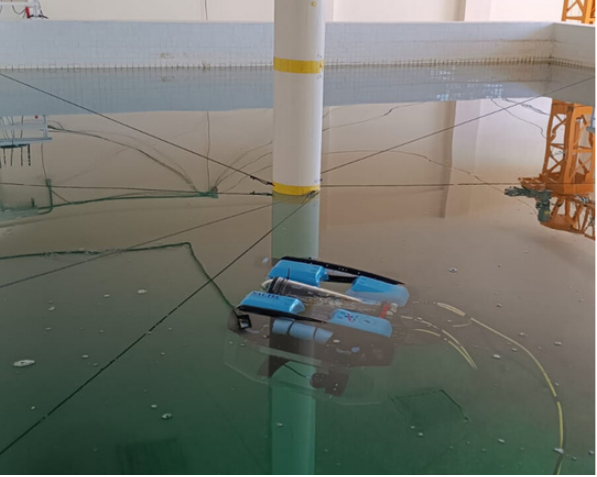}
        \subcaption{Experimental setup.}
        \label{subfig:real}
    \end{subfigure}
    \hfill
    \begin{subfigure}[t]{0.32\textwidth}
        \includegraphics[width=\linewidth]{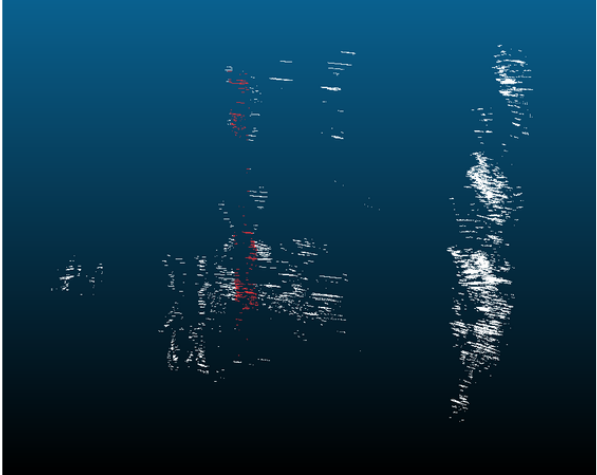}
        \subcaption{Reconstruction using ElevateNET.}
        \label{subfig:real-elevatenet}
    \end{subfigure}
    \hfill
    \begin{subfigure}[t]{0.32\textwidth}
        \includegraphics[width=\linewidth]{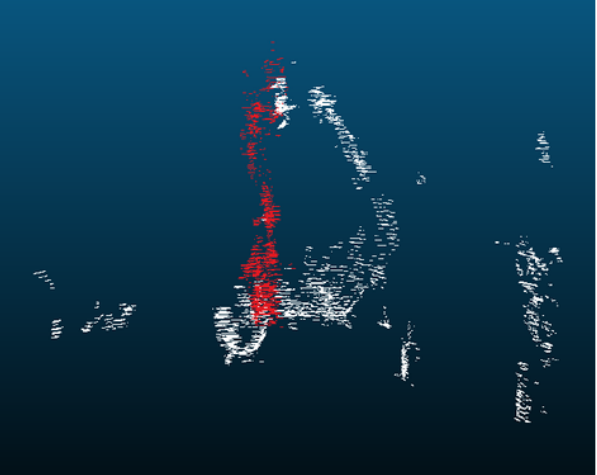}
        \subcaption{Reconstruction using ElevateNET-R.}
        \label{subfig:real-elevatenetR}
    \end{subfigure}
    \caption{Qualitative comparison of reconstruction results on real-world sonar data. \ref{subfig:real} The experimental setup, showing the plastic pipe target within the indoor tank. \ref{subfig:real-elevatenet} 3D reconstruction generated using the ElevateNET~\cite{debortoli2019elevatenet} methodlogy. \ref{subfig:real-elevatenetR} 3D reconstruction generated by the methodology proposed in this work. In both (\ref{subfig:real-elevatenet} and \ref{subfig:real-elevatenetR}), the red highlighted region indicates the segmented pipe, which is the primary object of interest.}
    \label{fig:RealDataRecosntruction}
\end{figure*}

In addition to the quantitative analysis, we conducted a qualitative evaluation on real-world data to assess the sim-to-real transfer performance. For this test, we selected the ElevateNET and the proposed ElevateNET-R models from our simulation experiments. These pre-trained models were then used to perform inference on the real sonar images. The resulting elevation maps were subsequently converted into 3D point clouds, providing a qualitative basis for comparison, as illustrated in the Figure~\ref{fig:RealDataRecosntruction}.

%% file: text/5_conclusion.tex
In conclusion, this paper presented the Synthetic Enclosed Echoes (SEE) dataset, a novel contribution to underwater 3D reconstruction using sonar data. The dataset includes 40 diverse objects, four scenarios, and detailed metadata, facilitating comprehensive evaluations of classical and learning-based reconstruction methodologies. Furthermore, this study proposed a modified ElevateNET architecture, demonstrating superior performance in 3D reconstruction compared to existing methods, even when generalized to unseen object data.

The proposed dataset and all code utilized in this study are publicly available \footnote{https://github.com/guilhermecdo/SEE}, facilitating community access and contribution. The simulation environment is designed to be expandable and maintainable, encouraging ongoing development and maintenance. The primary contribution of this work lies in the dataset and, particularly, the simulation environment, which enables the simulation of diverse sonar types and underwater structures.